Stylistic Clusters and the Syrian/South Syrian Tradition of First-Millennium BCE Levantine Ivory Carving: A Machine Learning Approach

Amy Rebecca Gansell[a,*], Jan-Willem van de Meent[b,d], Sakellarios Zairis[c], Chris H. Wiggins[d,c]

[a] The Department of Art and Design, St. John's Hall, Room B-11, 8000 Utopia Parkway, Queens, New York 11439 USA

[b] Department of Chemistry, Columbia University, 3000 Broadway, MC 3140, New York, New York 10027 USA

[c] Center for Computational Biology and Bioinformatics, Columbia University, 1130 St. Nicholas Avenue, New York, New York 10032 USA

[d] Department of Applied Physics and Applied Mathematics, The Fu Foundation School for Engineering and Applied Sciences, Columbia University, 500 West 120th Street, Mudd 200, MC 4701, New York, New York 10027 USA

[*] Corresponding Author. St. John's University, The Department of Art and Design, St. John's Hall, Room B-11, 8000 Utopia Parkway, Queens, New York 11439 USA.  Tel.: 001-617-669-2910; Fax: 001-718-990-2075; E-mail address: gansella@stjohns.edu (A.R. Gansell)

## Abstract

Thousands of first-millennium BCE ivory carvings have been excavated from Neo-Assyrian sites in Mesopotamia (primarily Nimrud, Khorsabad, and Arslan Tash) hundreds of miles from their Levantine production contexts. At present, their specific manufacture dates and workshop localities are unknown. Relying on subjective, visual methods, scholars have grappled with their classification and regional attribution for over a century. This study combines visual approaches with machine-learning techniques to offer data-driven perspectives on the classification and attribution of this early Iron Age corpus.

The study sample consisted of 162 sculptures of female figures. Among them are examples that have been conventionally attributed to three main regional carving traditions: "Phoenician," "North Syrian," "Syrian/South Syrian/Intermediate". We have developed an algorithm that clusters the ivories based on a combination of descriptive and anthropometric data. The resulting categories, which are based on purely statistical criteria, show good agreement with conventional art historical classifications, while revealing new perspectives, especially with regard to the contested "Syrian/South Syrian/Intermediate" tradition. Specifically, we have identified that objects of the Syrian/South Syrian/Intermediate tradition may be more closely related to Phoenician objects than to North Syrian objects; we offer a reconsideration of a subset of "Phoenician" objects, and we confirm Syrian/South Syrian/Intermediate stylistic subgroups that might distinguish networks of acquisition among the sites of Nimrud, Khorsabad, Arslan Tash and the Levant. We have also identified which features are most significant in our cluster assignments and might thereby be most diagnostic of regional carving traditions. In short, our study both corroborates traditional visual classification methods and demonstrates how machine-learning techniques may be employed to reveal complementary information not accessible through the exclusively visual analysis of an archaeological corpus.

## 1. Introduction and Background

Artifact classification is fundamental to archaeology, and statistical methods have long been used to typologize, seriate, and interpret objects (Baxter, 2008; Wilcock, 1999). In contrast, ancient art is more typically sorted by subjective assessments of style and iconography, with diagnostic features based on descriptive criteria that lack precise boundaries (Saragusti et al., 2005; Winter, 2005). However, since the 1970s, art historians have experimented with computer-based quantitative methods to classify figural sculpture according to proportional data and quantifiable features (Guralnick, 1973, 1976; Roaf, 1978). Recently Carter and Steinberg (2010) have revised Guralnick's pioneering contributions, while reasserting that statistical analysis can support and enrich interpretations formed by scholars on purely visual grounds.

This paper presents a data-driven approach to the characterization of groups of stylistically similar objects. Our methodology relies on well-established techniques from the machine learning community (Bishop, 2006) and groups objects based on purely statistical criteria without requiring art historical knowledge. Commonly employed in domains ranging from computer vision to neuroscience, machine learning methodologies have been successfully, but more rarely, applied to archaeological data (Boon et al., 2009; Menze et al., 2006; Nguifo et al., 1997; van der Maaten et al., 2007).

Engaging machine learning techniques to complement traditional art historical interpretations, this paper analyzes early first-millennium BCE Levantine ivory sculptures. At present, the artifacts' specific manufacture dates and production localities are unknown, as most were excavated from secondary, Neo-Assyrian contexts in Mesopotamia (Herrmann and Millard, 2003; Thomason, 2005). Only at the site of Samaria, in Israel, was a substantial assemblage discovered in the Levant itself, but this corpus, which is not fully published, was found in disturbed, Classical-period strata (Crowfoot and Crowfoot, 1938; Reisner, 1924; Suter, 2011; Tappy, 2006; Uehlinger, 2005).

Our goal is to evaluate art historically established classifications and to discover new information to aid in interpreting the objects' production origins and diffusion networks. We have developed a machine learning technique that clusters the ivories based on a combination of anthropometric and descriptive data. By elucidating relationships among a sample of female figures, our results contribute to over a century of efforts to classify the ivories and attribute their manufacture to various Levantine regions (Poulsen, 1912).

1.1 Geographical and Historical Setting

During the ninth and eighth centuries BCE, Levantine artisans created the objects under analysis out of African elephant tusks that had been imported to the Levant (Herrmann et al., 2009). "The Levant" encompasses the Mediterranean coast, much of modern Syria, and parts of southern

Anatolia. The northern Mediterranean coast of modern-day Israel, Lebanon, and Syria is referred to as "Phoenicia," and the interior is described as "North Syria" and "South Syria."

Reflecting their mobility, thousands of Levantine ivories were excavated from royal buildings at the Neo-Assyrian capital of Nimrud in Iraq (Barnett, 1957; Herrmann, 1986, 1992b; Herrmann et al., 2009; Mallowan and Herrmann, 1974; Orchard, 1967; Pappalardo, 2006, 2008, 2009; Safar and al-Iraqi, 1987). Much smaller assemblages were discovered at the Neo-Assyrian capital of Khorsabad in Iraq (Loud and Altman, 1938) and at the provincial Neo-Assyrian site of Arslan Tash in Syria (Cecchini, 2009; Thureau-Dangin et al., 1931). A smattering of additional ivories have been found at or linked to other Neo-Assyrian sites, including Nineveh and possibly Sharif Khan in Iraq (Barnett, 1957, 1982; Herrmann, 1986).

1.2 Levantine Ivory Sculpture

The ivory sculptures originally adorned and constituted furniture and luxury items. Their iconographic repertoire includes male and female figures, mythological and fantastical beings, animals, vegetation, and geometric designs. Based on West Semitic inscriptions on a few objects and visual comparisons to other Levantine media, most scholars recognize three Levantine carving traditions: "Phoenician," "North Syrian," and one that has variously been labeled "Syrian", "South Syrian", or "Intermediate" (Gubel, 2000; Herrmann, 1986, 1992a, 1992b, 2000, 2005; Herrmann et al., 2009; Millard, 1962; Scigliuzzo, 2005a; Wicke, 2005; Winter, 1976b, 1981, 1992, 1998b, 2005).

Phoenician works emulate Egyptian art (Fig. 1). They are also recognizable for their poised forms, spatially balanced tableaux, and openwork designs (Fig. 2).

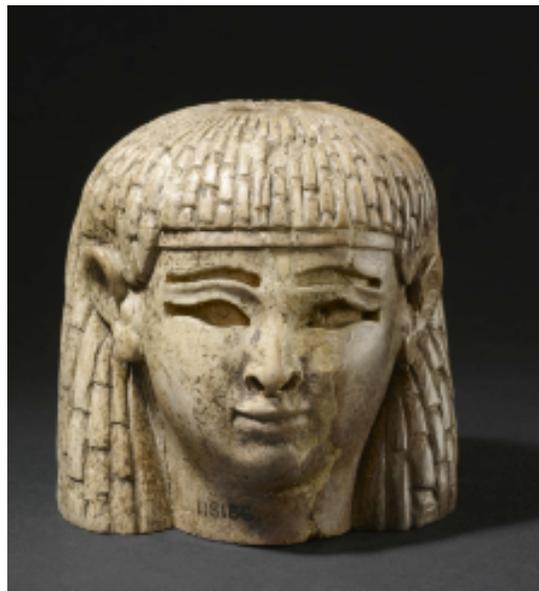

Figure 1: Female head, ivory. Nimrud, Iraq, c. 900—700 BCE. Ht. 6.1cm. [The British Museum, WA 118186, © Trustees of The British Museum]

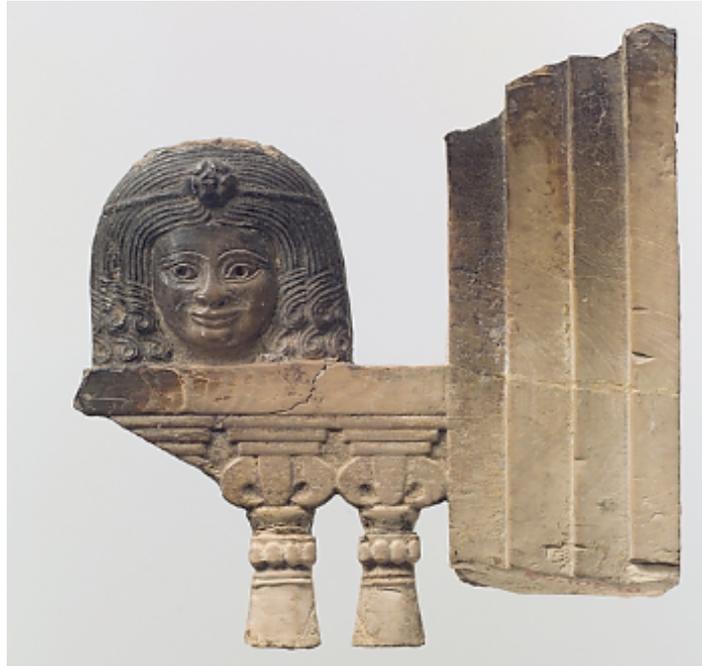

North Syrian figures are characterized by round faces, over-sized eyes, and/or stocky bodies (Figs. 3—5). Egyptian iconography is referenced, but North Syrian artists often corrupted motifs that were more accurately depicted in Phoenician art.

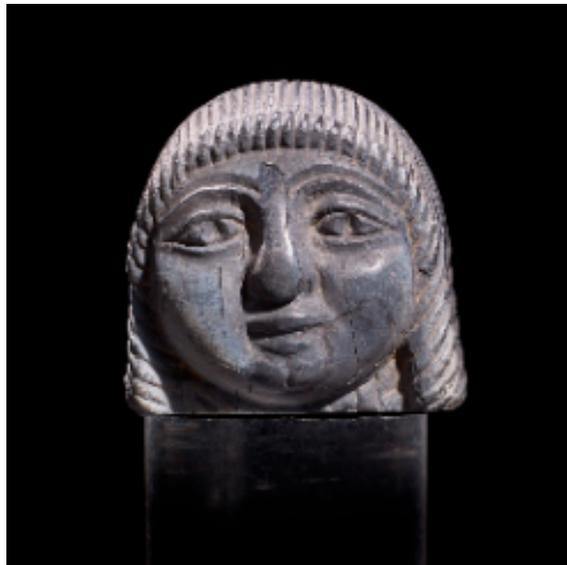

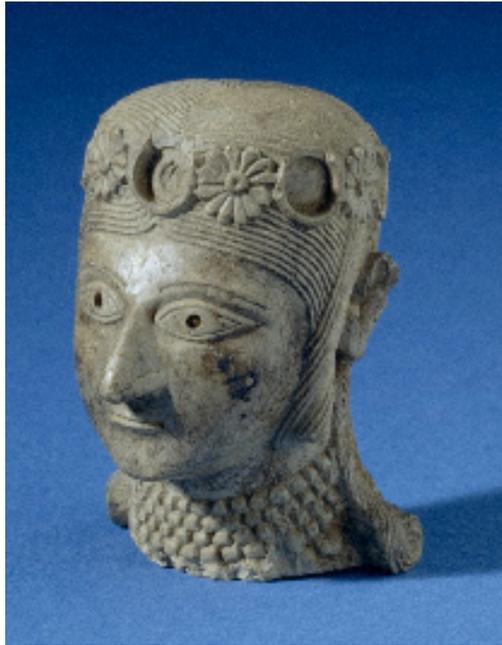

<u>Figure 4:</u> Female head, ivory. Nimrud, Iraq, c. 900—700 BCE. Ht. 4.3 cm. [The British Museum, WA 118234, © Trustees of The British Museum]

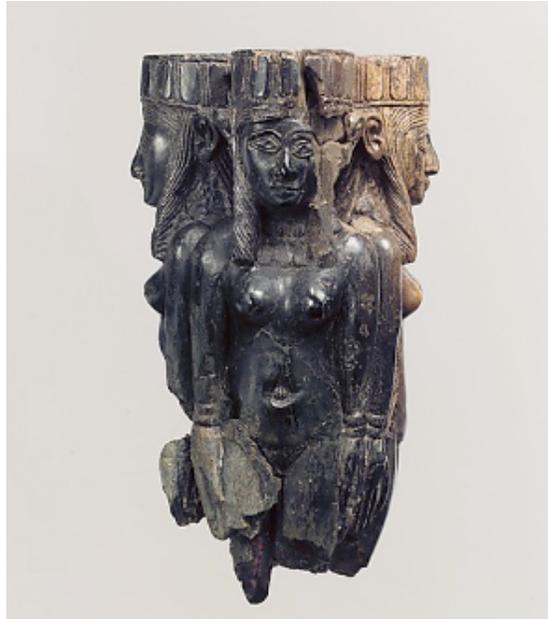

<u>Figure 5:</u> Handle or furniture support with four female figures, ivory and gold. Nimrud, Iraq, 8th century BCE. Ht. 9.91 cm. [The Metropolitan Museum of Art, Rogers Fund, 1952 (52.23.2). Photograph, all rights reserved, The Metropolitan Museum of Art. Image source: Art Resource, NY]

Finally, the Syrian/South Syrian tradition seems to be a hybrid of (or a tradition that is situated "between") the Phoenician and North Syrian traditions (Figs. 6—12).

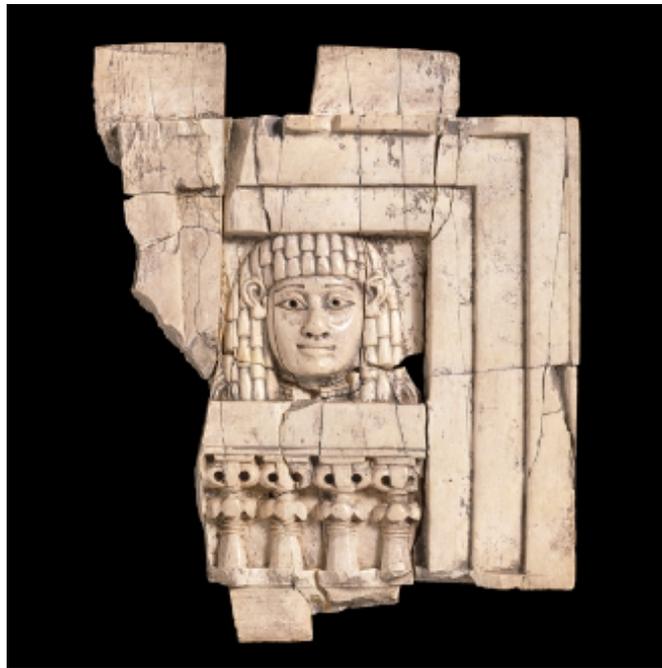

Figure 6: Woman-at-the-window plaque, ivory. Nimrud, Iraq, 900—700 BCE. Ht. 11 cm. [The British Museum, WA 118159, © Trustees of The British Museum]

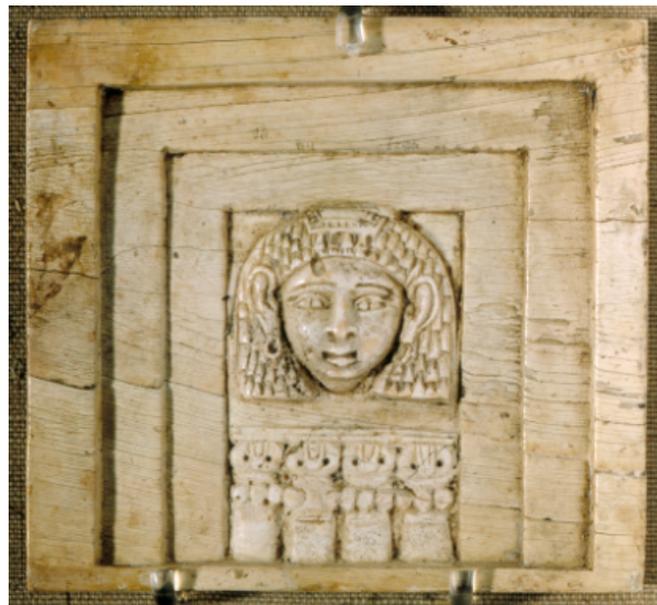

Figure 7: Woman-at-the-window plaque; ivory, formerly gilded. Arslan Tash, Syria, 8th century BCE. Ht. 8.1 cm. [Louvre, AO 11459. Réunion des Musées Nationaux / Art Resource, NY]

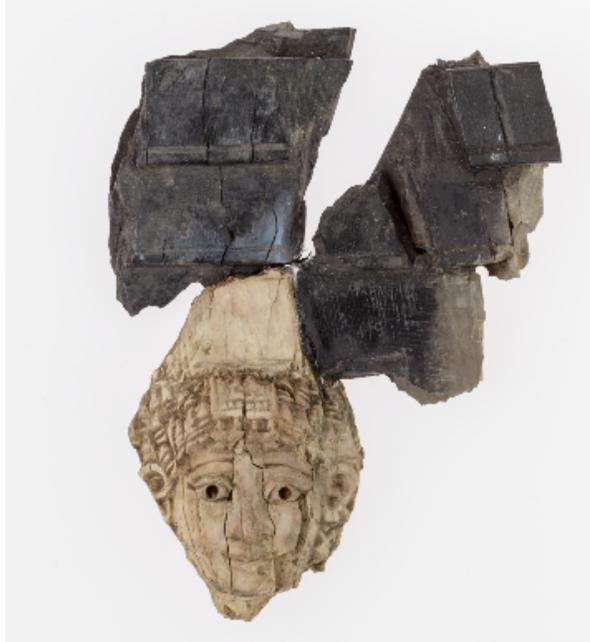

Figure 8: Woman-at-the-window plaque, ivory. Khorsabad, Iraq, 8th century BCE. Ht. of face fragment 5.7 cm. [Oriental Institute of the University of Chicago, A 22164. Courtesy of the Oriental Institute of the University of Chicago. Photo: Anna R. Ressman]

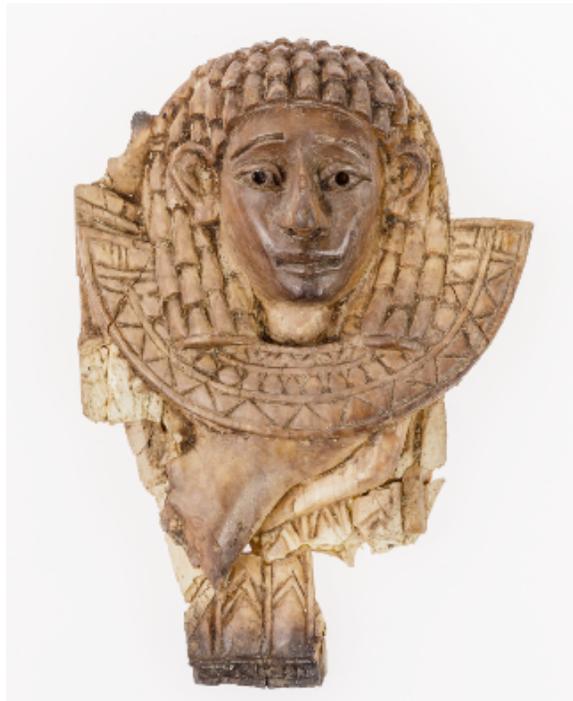

Figure 9: Female-headed sphinx, ivory. Khorsabad, Iraq, 8th century BCE. Ht. of face fragment 5.8 cm. [Oriental Institute of the University of Chicago, A 22172. Courtesy of the Oriental Institute of the University of Chicago. Photo: Anna R. Ressman]

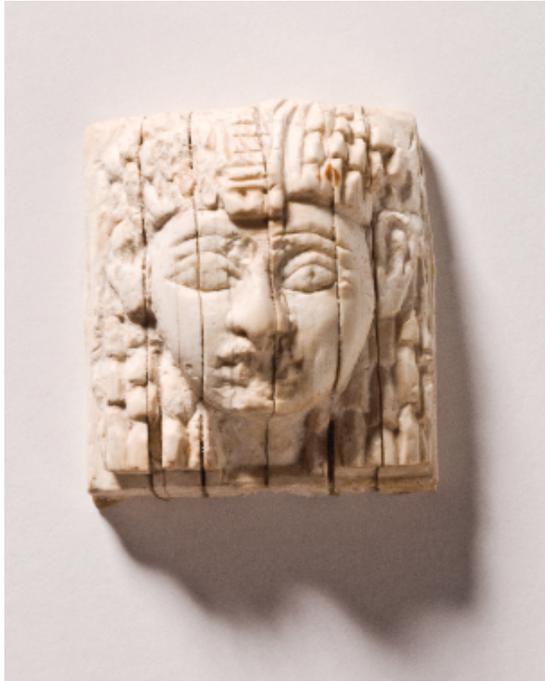

Figure 10: Female face, ivory. Arslan Tash, Syria, 8th century BCE. Ht. 2.5 cm. [Badisches Landesmuseum Karlsruhe, 72_35]

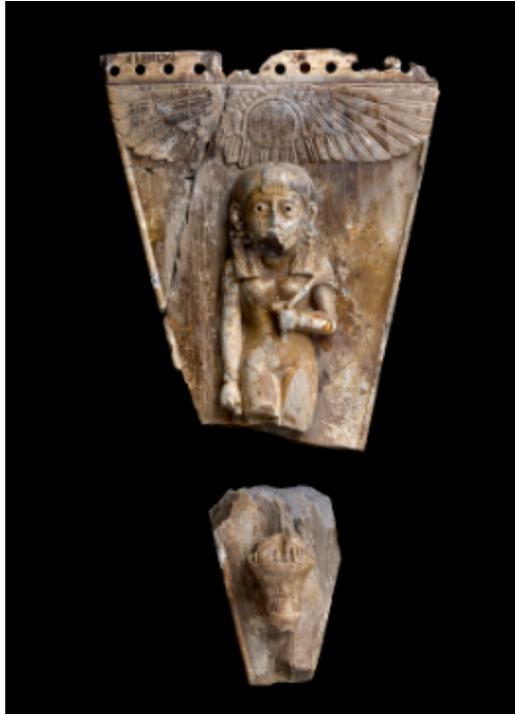

Figure 11: Plaque depicting a nude female figure, ivory. Nimrud, Iraq, 9th—8th century BCE. Ht. 10.16 cm. [The British Museum, WA 118104, © Trustees of The British Museum]

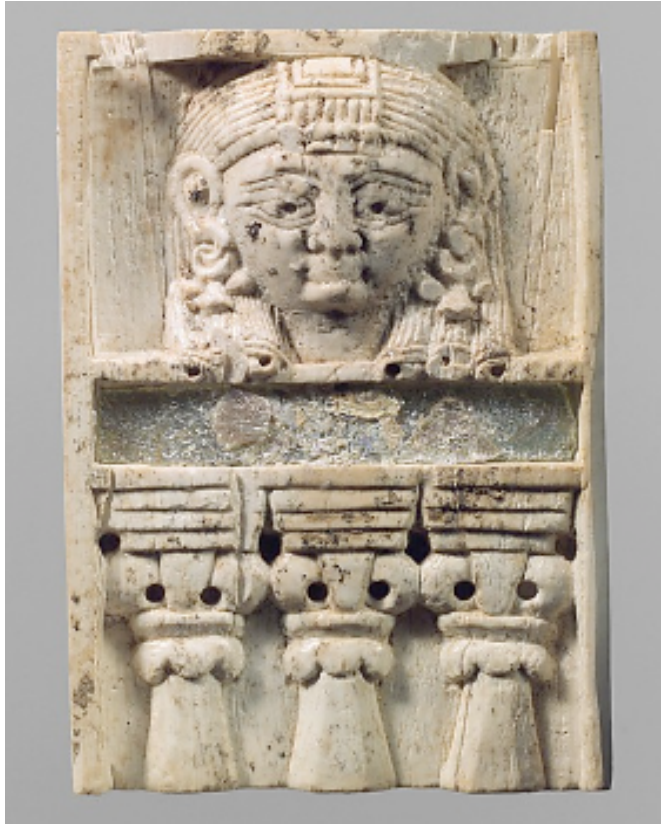

Figure 12: Woman-at-the-window plaque, ivory and glass. Arslan Tash, Syria, 9th—8th century BCE. Ht. 6.4 cm. [The Metropolitan Museum of Art, Fletcher Fund, 1957 (57.80.12). Photograph, all rights reserved, The Metropolitan Museum of Art. Image source: Art Resource, NY]

Within the traditions, which themselves are debated, art historians have proposed, but not agreed upon, "style-groups" that could differentiate dates of production and regional workshops, and/or reflect the "hands" of various artists (Herrmann, 1989, 1992a, 2000, 2005; Herrmann et al., 2009; Scigliuzzo, 2005a, 2005b, 2006, 2009; Suter, 1992; Winter, 1976a, 1981, 1992, 1998b, 2005). Style-groups are typically defined by the manner of rendering motifs, such as trees, animals, and human figures. Most of the established criteria for recognizing style-groups derive from non-figural iconography, but scholars have acknowledged and demonstrated the potential of sorting the ivories according to facial features and physical proportions (Scigliuzzi, 2009; Suter, 1992; Wicke, 2009; Winter, 1976a, 1981, 2005).

The Phoenician tradition includes the "Egyptianizing", "Ornate", and other style-groups; however, the stylistic classification of the Phoenician works in our sample is tenuous (Figs. 1, 2). The North Syrian tradition contains, among others, the "Round-Cheeked and Ringletted" group (Fig. 3) and the "Flame and Frond" group (Affanni, 2009; Herrmann, 1989, 1992a, 1992b, 2002; Herrmann et al., 2009; Wicke, 2005) (Figs. 4, 5). Perhaps the broadest style-group is the Syrian/South Syrian "Wig and Wing" group (Herrmann, 1992a, 1992b; Scigliuzzo, 2005b, 2009) (Figs. 6—10). No matter how the traditions and their style-groups are structured, some ivories seem to fall between established designations, and others demonstrate relationships to more than one group (Brown, 1958; Herrmann, 1992b, 2005; Herrmann et al., 2009; Winter, 1981, 2005).

## 2. Materials and Data

2.1 The Research Sample

We focus on female figures (including female-headed sphinxes), because we had available a dataset from a previous project (Gansell, 2008, 2009). The original dataset included 210 sculptures, representing almost every Levantine ivory sculpture portraying a female figure held in a reasonably accessible museum collection in the United States, Europe, and Syria. We culled this sample by removing artifacts that did not offer data for the variables considered here or that were physically configured according to different proportional standards because they portrayed figures in profile only. A few other artifacts (such as stone carvings) initially included as test cases were removed from the sample. Our final sample included 162 objects, comprising 126 sculptures from Nimrud, seven from Khorsabad, and 26 from Arslan Tash (Appendix A). In addition, one ivory is believed to be from Nineveh, and two were allegedly found at the site of Sharif Khan. There are 14 female-headed sphinxes (Fig. 9), 32 woman-at-the-window plaques (Figs. 2, 6—8, 12), 41 whole and fragmentary full-length figures (Figs. 5, 11), and 75 heads and faces (Figs. 1, 3, 4, 10). Faces (measured from chin to hairline) range from 11.4 cm to 0.45 cm in height; most are in the smaller end of this range.

2.2 Data Collection and Preparation

All data were collected firsthand by taking caliper measurements and recording iconographic and stylistic characteristics ("categorical" data) using a standard questionnaire (Gansell, 2008). While numeric, anthropometric values define ratios, such as the thickness versus the length of the lips, categorical input describes features, such as the "shape" of the lips, that cannot be measured in real-valued continuous quantities. Since a previous computational analysis of the categorical data indicated that features relating to the eyes and hair were strongly associated with production source (Gansell et al., 2007), we edited the original dataset to incorporate only features preserved from the shoulders up. Limiting the number of variables also served to produce a dataset more amenable to statistical analysis by decreasing the percentage of missing data in the sample.

The final dataset was reduced from 137 to 33 features. Twenty-one features are categorical, and 12 are anthropometric. The anthropometric features include 11 ratios derived by converting raw anthropometric measurements into facial proportions and one numeric feature, the pupil diameter, which was kept un-normalized, because this could document the types of tools used (Herrmann, 1986) (Fig. 13).

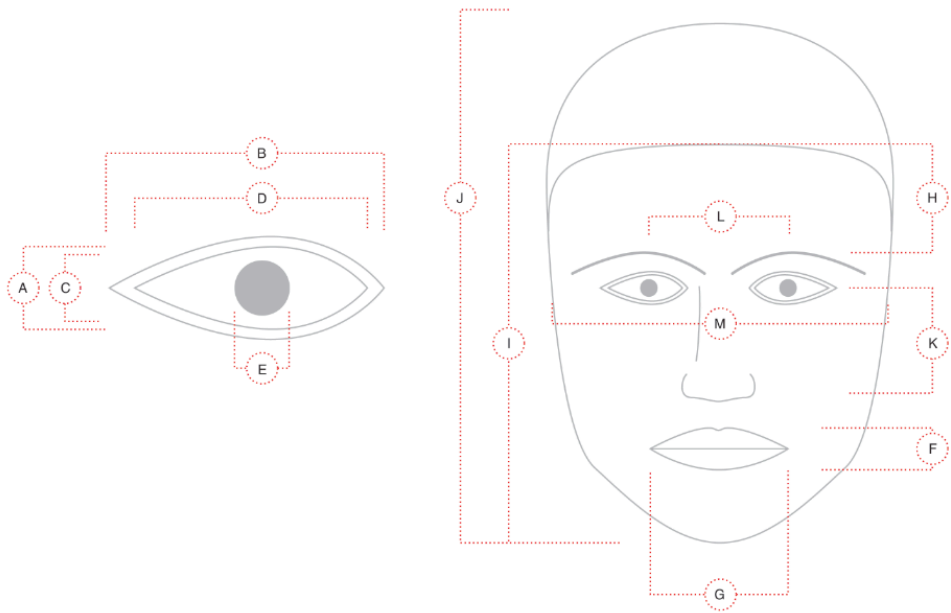

Figure 13: Diagrams of anthropometric features (listed in Appendix B), calculated as follows: 1 = a/b, 2 = c/d, 3 = c/a, 4 = e/c, 5 = f/g, 6 = h/i, 7 = i/j, 8 = a/i, 9 = k/i, 10 = l/i, 11 = l/m, 12 = e. [Illustration: Jasper Pope]

2.3 Sparsity of Data

Many of the ivories are fragmentary and/or suffer surface damage. Such sculptures did not allow for all features to be recorded. Although we constructed our set of features with the aim of minimizing missing data, the final tabulated data still contained roughly 37% missing values. We discuss in Section 3.1 how this issue of sparsity motivates our choice of statistical model.

**3. Methodology and Analysis**

Just as an expert might struggle to classify a sculpture, an expressive model should maintain probabilities of belonging to different clusters rather than a single, hard assignment to a group. Mixture models are a natural choice for such "soft" assignment problems (Bishop, 2006). In a mixture model, the range of anthropometric and categorical features found within each stylistic group is described by a probability distribution. Each sculpture in a group is assumed to be an independent and identically distributed (IID) sample from this distribution. The mean of the distribution defines the "stylistic center" for the group, whereas the standard deviation represents the "stylistic variation." An "expectation-maximization" (EM) procedure may then be used to determine how likely it is that each object belongs to a given group, as well the means and variances of its features. This "unsupervised" method offers the advantage of not requiring any explicit annotation, such as the art historically interpreted carving tradition.

In order for mixture model approaches to be effective, each group must be characterized by a sufficiently distinct set of feature values. In our analysis, it is not a priori apparent which features will predict stylistic tradition. For this reason, we use a model that characterizes each cluster by a combination of anthropometric and categorical features. There is a precedent for applying this method, sometimes called a "mixed-mode" approach, to archaeological samples (Baxter et al., 2008; Dellaportas 1998).

Both feature sets (anthropometric and categorical) have potential advantages and disadvantages. Anthropometric features, such as the size and spacing of the eyes, may contain valuable information, but, because the objects are different sizes, we must select some base length, such as the length of the face. There is no guarantee that the resulting ratios directly report on stylistic choices. Categorical features more explicitly capture artistic decisions, such as the type of hairstyle depicted, but there is a risk that the chosen categories implicitly reflect the observer's pre-existing notions. Nonetheless, stylistic form is ultimately a subjective notion that may not be readily captured in numbers alone (Winter, 1998a); thus, we expect categorical features to play a role in determining cluster membership. From a statistical perspective there is a question of bias, but since our goal is to construct a quantitative statistical description that can be compared to existing art historical interpretations, it is not necessarily undesirable to perform analysis on features that correlate with such notions.

### 3.1 Mathematical Model of the Data

There are two important properties of the data to consider when performing mixture model analysis for this corpus. First, the number of features (33) is relatively large compared to the number of objects (162). Recent work on "spectral" clustering methods suggests higher dimensional data can be easier to separate (Hsu and Kakade, 2013), but local-maxima problems in EM approaches are exacerbated in high-dimensional data. Analysis results are also more difficult to interpret. The second and more challenging characteristic of this dataset is that 37% of the feature values are missing. Numerous strategies can be formulated for dealing with missing data in models where there is a conditional dependence on missing values (Gharahmani and Jordan, 1995). We sidestep the need for such approaches by assuming that all features are independent. A mixture model with such uncorrelated features has a likelihood

$$p(x, y \mid \theta) = \sum_z p(x, y \mid z, \theta) = \prod_n \sum_{z_n} \prod_k \left[ p(x_n \mid \theta_k) p(y_n \mid \theta_k) p(z_n) \right]^{z_{nk}},$$

$$p(y_n \mid \theta_k) = \prod_c \text{Categ}\left( y_n^c \mid \rho_k^c \right) = \prod_d \prod_c \left( \rho_k^c \right)^{y_{nd}^c},$$

$$p(z_n) = \prod_k \pi_k^{z_{nk}}$$

where the variables *x, y,* and *z* represent the anthropometric features, categorical features, and cluster membership respectively

$\{x_n^r\}$ : data for object *n*, anthropometric feature *r*

$\{y_n^c\}$ : data for object *n*, categorical feature *c*

$\{z_n\}$ : cluster membership for object *n*

and the model parameters (collectively referred to as $q$) are given by

$\{\pi_k\}$: prior probability for membership to cluster $k$

$\{\mu_k^r\}$: observation mean for anthropometric feature $r$ in cluster $k$

$\{\lambda_k^r\}$: observation precision (1/variance) for anthropometric feature $r$ in cluster $k$

$\{\rho_k^c\}$: observation probability for categorical feature $c$ in prototype $k$.

A benefit of this model is computational simplicity. Missing values can be integrated out trivially and therefore do no contribute to the likelihood in EM estimation. However, our main motivation for using this simplest model is that it requires fewer parameters. For example, a model with K states requires $KR^2$ parameters to express a full covariance on R anthropometric features. For a dataset of N objects with average sparsity σ, this means there are σN/KR. For this corpus we would have 8.5/K observables per parameter, which raises significant concerns for overfitting. Consequently, we consider this simplest model the most appropriate for a first-pass clustering analysis, given the small size and sparsity of the corpus.

3.2 Inference Procedure

The mixture model defined above describes each stylistic group according to a set of model parameters $q$. These parameters in turn characterize the range of anthropometric and categorical features for sculptures in each group. The objective of our modeling approach is to automatically determine the most appropriate set of stylistic parameters associated with each group, as well as the group assignment of each sculpture.

Maximum likelihood estimation is a standard technique that may be used to infer these stylistic parameters in an unsupervised manner (Bishop, 2006). The underlying idea is to find the set of parameters that is most likely to yield the set of features seen in our corpus. A recognized problem of maximum likelihood methods is that estimating numeric parameters yields ill-defined results when a group contains only a single object. This causes an apparent zero variance, or infinite precision. As a safeguard against such numerical instabilities, we regularize this model by introducing a Gamma prior on the feature precision (Bishop, 2006):

$$\lambda_k^r \sim \text{Gamma}\left(a_k^r, b_k^r\right).$$

This probability specifies what range of feature variances we expect to see in the data and produces a model that treats unusually small or large variances as highly unlikely. The hyperparameters for this prior are chosen in such a manner as to prevent numerical instability, while, in most cases, allowing our likelihood to overwhelm the prior's influence. This form of inference is called maximum a posteriori (MAP) estimation (Bishop, 2006) and is concerned with maximizing the *product* of the likelihood and the prior, rather than just the likelihood – which is equivalent to maximizing the regularized log-likelihood

$$L^{\text{MAP}} = \log\left[p(x, y \mid \theta)p(\theta \mid a, b)\right] = \log\left[p(x, y \mid \theta)\right] + \log\left[p(\theta \mid a, b)\right].$$

The specific inference scheme employed for this optimization task is the well-known EM algorithm (Bishop, 2006), wherein parameter update rules are iteratively applied until the log-likelihood value converges to some maximal value (Dempster et al., 1977). The parameter settings at this maximal value of the log-likelihood specify the model learned from the data.

A common concern in EM inference is convergence upon parameter settings that are only *locally* optimal and do not represent the best *global* setting of the parameters (Bishop, 2006). To remedy this we perform 500 restarts, each with a randomly initialized set of parameters, and we keep only the best performer as our final model.

Our source code is freely available for use at http://aneic.github.com.

3.3 Performance Metrics & The Oracle

A recurring question in mixture model analysis is how many clusters to fit to the data. This problem is known as model selection, and one standard procedure for determining the optimal number of clusters in this setting is cross validation (Bishop, 2006). In cross validation the data are partitioned into distinct subsets, and the model is trained on a different subset of the overall data from that which is used to evaluate the model's performance. This approach is known to suffer, however, when the dimensionality of the data is high relative to size of the dataset (Smyth, 2000). We implemented five-fold cross validation but did not observe any peaking behavior of the log likelihood as a function of K, the number of clusters. We did observe that our posterior distribution was peaked for the range of priors under consideration.

Because the log posterior in itself should not be used as a model selection criterion, we have developed an additional indicator: a set of hand-curated pairings of ivories that should cluster together under any reasonable model. This "oracle", as we call it, encodes archaeological common sense, such as demanding that two figures depicted on the same object should be found in the same cluster (Fig. 5). We quantify agreement of our clusters against this oracle and, again, see peaking behavior in the same range of K-values as the posterior distribution (Fig. 14).

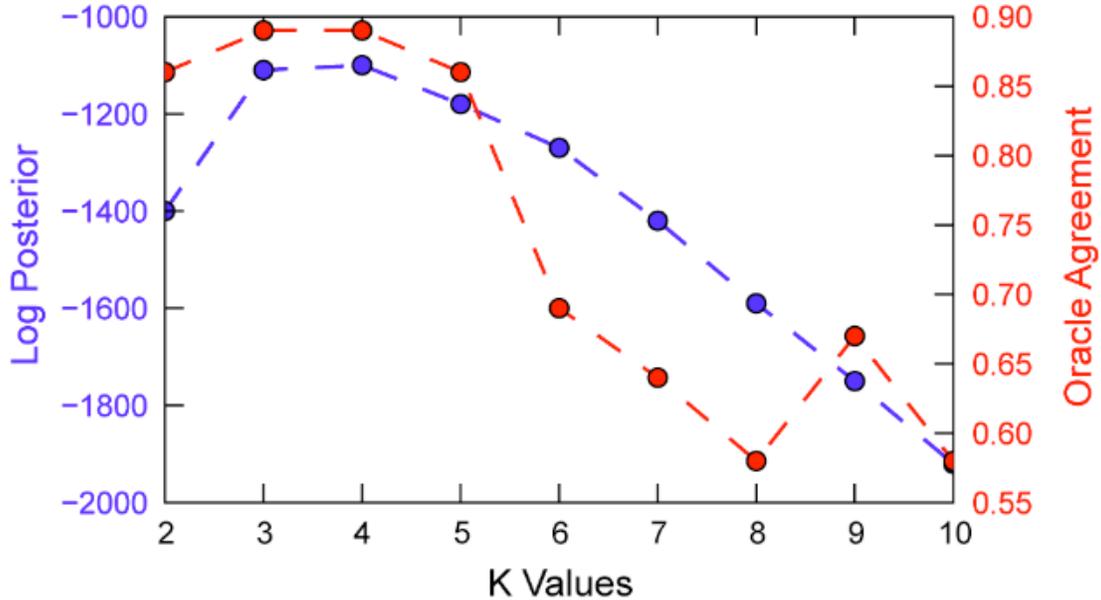

Figure 14: Blue shows the log posterior of the model, and red plots the agreement of the clustering with a predefined set of sculpture pairings, termed the "oracle".

On the graph, there is a close margin between the peak performance of K=3 and K=4, but the cluster content at K=4 is more consistent with art historical domain knowledge (see Section 4.1). While K=3 performed "well" by not separating related objects, it also grouped unrelated or too broadly related objects.

### 3.4 Network Visualization of Dataset

In order to better comprehend the clusters produced by our algorithm we use graph-visualization techniques commonly engaged in the study of social networks. We first construct a 162 x 162 matrix where the elements are the distances between pairs of ivories. The distance metric is the sum of square differences across all anthropometric features and attributed a flat penalty for non-matching categorical features:

$$D_{ij} = \sum_r \left( x_i^r - x_j^r \right)^2 + \sum_c \left( y_i^c \neq y_j^c \right) / 20 \, .$$

Note that missing values are not included in the summation, and, as a result, sparser objects appear closer. Any two objects whose distance matrix element was less than 0.5 satisfied our criterion for adjacency. We only draw edges between objects in the network below if they are adjacent, so 56 points that were not found to be adjacent to any other are not represented. A force-directed graph layout was used to embed the network in two dimensions (Fruchterman and Reingold, 1991) (Fig. 15).

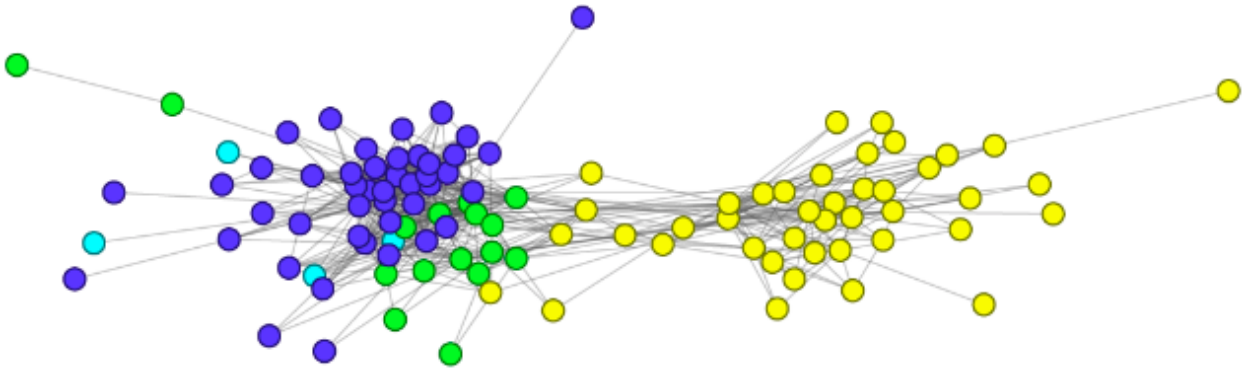

Figure 15: A subset of ivories from the larger corpus is represented as connected nodes in a graph with the results of our clustering algorithm color-coded. Yellow: Cluster 1, Blue: Cluster 2, Cyan: Cluster 3, Green: Cluster 4.

Although this visualization does not present a precise quantitative analysis, the large degree of agreement with the probabilistic partitioning of the data produced by our algorithm is striking, since the network illustrated in Figure 15 did not incorporate any knowledge of these clustering results. This agreement suggests that the algorithm is indeed assigning cluster membership based on some notion of distance in the high dimensional feature space.

## 4. Results and Discussion

Our results corroborate art historical classifications and yield new perspectives not accessible through visual analysis alone. First, we compare the clustering results with art historical classifications to demonstrate their commonalities at K=4. Second, an interpretation of the clusters at K=4 sheds light on the membership and production origins of the Syrian/South Syrian tradition. Third, and, finally, we are able to report which features carry the most weight in cluster assignment at K=4.

### 4.1 The Clusters

There is no comprehensive "key" indicating the ivories' art historically assigned traditions and style-groups, but, based on published examples, Appendix A presents the most reasonable art historical classification for 150 out of 162 objects, and Figure 16 visualizes the distribution of ivories across clusters according to art historical tradition and site of discovery.

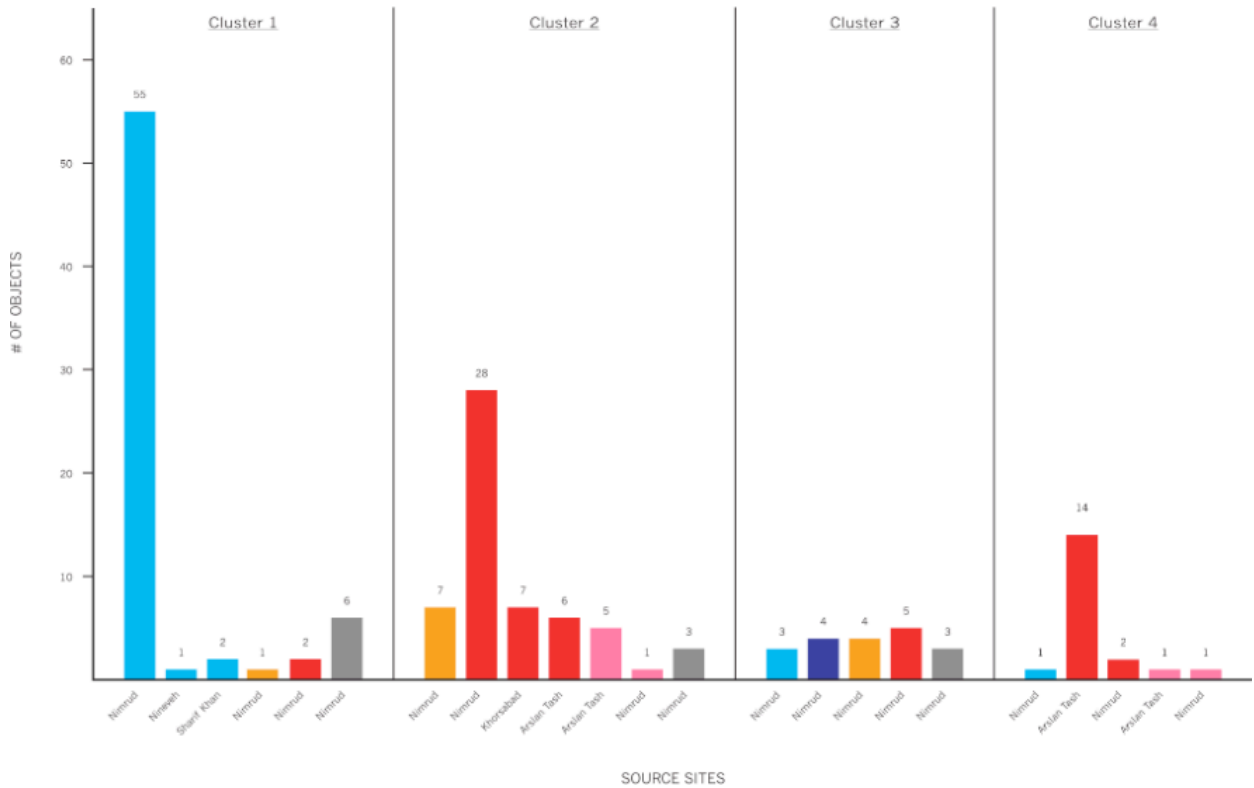

Figure 16: Quantification by cluster (at K=4) of objects by tradition/style-group from each site. Light Blue: North Syrian Flame and Frond, Dark Blue: North Syrian Round-Cheeked and Ringletted, Orange: Phoenician, Red: Syrian/South Syrian Wig and Wing, Pink: Syrian/South Syrian of unclassified style, Gray: Unclassified. [Illustration: Jasper Pope]

As indicated above, we focus on the results produced at K=4 (http://aneic.github.io/results/130317_01/130317_01_K04.html), because the output at K=3 (http://aneic.github.io/results/130317_01/130317_01_K03.html) is less coherent; for example, one cluster contains a mixture of 80 sculptures from all traditions. In comparison, K=4 discovers groups of 67 (Cluster 1), 57 (Cluster 2), 19 (Cluster 3), and 19 (Cluster 4) objects, generally organized as follows: Flame and Frond North Syrian objects (Cluster 1), Phoenician works with Wig and Wing Syrian/South Syrian sculptures (Cluster 2), a small but potentially meaningful mixture (Cluster 3), and Wig and Wing Syrian/South Syrian ivories, mostly from Arslan Tash (Cluster 4). These four clusters do not mimic the three art historical classifications, but they do largely maintain art historical groupings at the level of style-group.

4.2 The Syrian/South Syrian Tradition

The Syrian/South Syrian tradition was previously termed "Intermediate" on the grounds that the objects' hybrid appearance is not necessarily tied to their geographic origin (Herrmann, 1986, 2005; Herrmann et al., 2009; Scigliuzzo, 2005a). Otherwise, scholars have and continue to designate the works "Syrian" or "South Syrian," in reference to their hypothetical production region (Gubel, 2000; Herrmann et al., 2009; Winter, 1981, 1992). Less culturally and politically unified than Phoenicia and North Syria during the early first millennium BCE, Syria/South Syria encompassed Aram, Israel, and Judah. Conceivably participating in a pan-Near Eastern culture of exotic materials and iconography that affirmed elite identity and ideology, each kingdom could have had its own ivory-working center/s to supply royal consumption (Suter, 2011; Winter, 1981).

Most scholars concur that the ivories from Arslan Tash and Khorsabad, many from Nimrud, and most of the Samaria corpus belong to the Syrian/South Syrian tradition (Herrmann, 1986, 1992, 2005; Herrmann et al., 2009; Millard, 1962; Scigliuzzo, 2005a, 2009; Suter, 2011; Wicke, 2009; Winter, 1981, 1992). Recently, however, Scigliuzzo (2005b, 2006, 2009) has questioned the tradition's actuality, and Wicke (2009) has proposed that at least some "Syrian/South Syrian" sculptures could be lower quality Phoenician carvings.

Our clusters strengthen the possibility that Syrian/South Syrian objects are at least more closely related to Phoenician, rather than North Syrian, ivories (Scigliuzzo, 2005b; Wicke, 2009). Cluster 1 contains 58 of the sample's 62 North Syrian sculptures, while Clusters 2, 3, and 4 contain 70 of the 72 Syrian/South Syrian objects and all 11 Phoenician works (see Figs. 15, 16). If the clustering results are diagnostic of geographic proximity, this would also place their origin closer to Phoenicia (i.e., in South Syria).

Next, Cluster 3 at first appears to be a catchall, but its mixed nature encourages a reconsideration of "Phoenician" woman-at-the-window plaques. Along with three unclassified works, this cluster includes five Syrian/South Syrian objects, seven North Syrian works, and four Phoenician ivories. Three of the four Phoenician sculptures depict the only "Phoenician" woman-at-the-window figures in our sample (Fig. 2). These works, then, might be more closely associated with the Syrian/South Syrian tradition and/or the North Syrian tradition, than with other Phoenician images. Thus, they could be a Phoenician subtype influenced by Syrian/South Syrian and/or North Syrian style, a North Syrian style emulating the Phoenician tradition, or, reflecting their "in between" status, they could represent a Syrian/South Syrian subgroup.

Finally, cutting across sites, our clusters confirm Syrian/South Syrian subgroups that could be the legacy of multiple geographical, chronological, commercial, or artistic sources. This implies different political or mercantile relationships among Levantine production sites and Neo-Assyrian consumers, and it could point to the redistribution of Levantine ivories among Neo-Assyrian sites (Cecchini, 2009; Herrmann, 1986, 1992b; Herrmann et al., 2009; Mazzoni, 2009; Winter, 1976b, 1982, 1989, 1992, 1998b, 2005).

Cluster 4, for example, demonstrates a strong tendency for the Syrian/South Syrian Arslan Tash objects depicting figures wearing Egyptian-style wigs (Figs. 7, 10) to cluster independent of Syrian/South Syrian ivories from other sites. Despite the small sample size, this pattern suggests that they belong to a Syrian/South Syrian subgroup that is more distantly related to other Syrian/South Syrian objects. In Cluster 2, the concentration of five of the six Syrian/South Syrian figures with straight hair ending in curls (in the manner of Fig. 12) implies another subgroup. Additionally, the occurrence of all seven Khorsabad ivories in Cluster 2 supports their unified identity as Syrian/South Syrian subgroup, to which some nearly identical objects from Nimrud and Samaria not in this sample also belong (Cecchini, 2009; Herrmann, 1986; Scigliuzzo, 2009; Suter, 2011; Winter 1981).

4.3 Most Important Features

Art historians have long debated how best to sort Levantine ivory sculptures, with some approaches relying on an individual diagnostic feature or a pair of diagnostics (Herrmann, 1986; Herrmann et al., 2009; Scigliuzzo, 2009; Winter, 1992, 1998b, 2005). A benefit of our analysis is that it reveals and ranks the most predictive features in the formation of a cluster, quantitatively demonstrating which batch of anthropometric and categorical features might bear the most weight in grouping the ivories.

To assess the predictiveness of individual features, we calculate a quantity known as "mutual information" (Bishop, 2006), which is a measure of the agreement between feature values and cluster assignments. The most significant categorical features proved to be the shape of the hairline followed, in ranked order, by the form of the eyes, headdress type, the manner in which the nostrils were carved, hairstyle, and the presence/absence of a necklace.
The impact of anthropometric data is somewhat more difficult to assess. Numeric values must first be binned to calculate the mutual information, and the resulting ranking is not independent of the number of bins used. Moreover, features are not consistently ranked in the same order under different splits of the data. For example, in the first of five folds of the iteration producing four clusters, the ratio of lip thickness to lip length emerged as the third most significant variable, and in the fourth of these five folds, the mean pupil diameter emerged as the third most significant variable. Yet, the average rank of these anthropometric features is 15, as seen in Table 1.

| Feature | Average Rank | Data Type | Sparsity |
|---|---|---|---|
| shape of hairline | 1 | categorical | 0.24 |
| form of eyes | 2 | categorical | 0.23 |
| headdress | 3 | categorical | 0.20 |
| nostril carving technique | 4 | categorical | 0.54 |
| hairstyle | 5 | categorical | 0.20 |
| necklace | 10 | categorical | 0.47 |
| lip thickness : lip length | 15 | numeric | 0.46 |
| mean pupil diameter | 15 | numeric | 0.38 |

Table 1: Ranking of features in terms of the mutual information between cluster assignment and feature values. Average feature rank was computed from the harmonic mean of all instances. The average sparsity across the dataset is 0.37 (+/- 0.13), whereas the average sparsity for features listed in this table is 0.34 (+/- 0.13).

The above results correspond to a preliminary mutual-information-based analysis of the categorical data demonstrating that among the most significant features in the determination of carving tradition are eyes, the manner in which the nostrils were carved, and hairstyle (Gansell 2008, Gansell et al., 2007). Across ancient Near Eastern art, eyes and hair are consistently rendered in detail, and ancient Near Eastern literature praises their beauty (Gansell, 2008, in press; Winter, 2005).

**5. Summary, Future Directions, and Conclusion**

This analysis shows that statistical techniques can effectively be used to build and test our intuition when identifying stylistically similar objects within archaeological assemblages, even when there is a high degree of missing data. Our methodology, which uses purely statistical criteria and incorporates no knowledge of existing notions regarding the resemblance of objects, recovers a group structure that intersects with art historical hypotheses, especially with regard to the Syrian/South Syrian carving tradition. Beyond mere classification, our clustering results imply ancient Near Eastern cultural, social, historical, commercial, and political relationships that shed light on the ivories' regional production and the dynamic networks connecting Levantine kingdoms and the Neo-Assyrian empire. In addition, by revealing and ranking the most predictive features in the formation of a cluster, our results could illuminate ancient aesthetic preferences.

The model with independent features employed here can be thought of as the simplest possible mixture model that describes data with a combination of anthropometric and categorical features. Given the difficulty in determining the number of clusters for high-dimensional sparse data, a potential follow-up to this work could be to consider a fully Bayesian model with a Dirichlet Process prior where inference may be performed with either variational (Blei and Jordan, 2006) or sampling methods (Neal, 2000). Such "non-parametric" methods offer the promise of eliminating the need for cross-validation by determining the number of states required as part of inference procedure.

Eventually we hope to expand our sample beyond the 162 objects analyzed here. The Iraq Museum ivories, which are currently inaccessible, would offer hundreds of sculptures that fit with this corpus, while new excavations could reveal fresh assemblages at any time. The Samaria ivories, once fully published, would be especially informative to incorporate. It would also be productive to consider iconographic varieties of ivory sculpture other than female figures, and/or to look at makers' marks, fitters' marks, and variables of technical craftsmanship (Herrmann et al., 2009; Millard, 1962, 1986, 2005; Scigliuzzo, 2009).

Future research could combine the ivory evidence with other material sources, such as a corpus of metal bowls bearing West Semitic inscriptions and Levantine-style engravings (Barnett, 1967; Gubel, 2000; Markoe, 1985; Wicke et al., 2010; Winter 1981). Integrating additional evidence would not only help to refine the classification and attribution of the ivories, but it could also enhance our understanding of ancient Near Eastern spatio-temporal networks (Herrmann, 1986, 1992a; Suter, 2011; Uehlinger, 2005; Winter, 1976b, 1981). Ideally, we would recommend the application of our results in a broader social network analysis investigation (Knappett, 2011, 2013).

We conclude that this project represents a productive collaboration between science and the humanities. The field of archaeology is being transformed through opportunities to confront questions of history and culture with the powerful computational tools of science. Applying analytical technologies to archaeological data, machine learning has established itself to be as much a device for discovery as the pick and trowel.

**Acknowledgments**

The following museums generously granted access to ivories in their collections: The Aleppo National Museum; The Ashmolean Museum; Das Badisches Landesmuseum, Karlsruhe; Birmingham Museums and Art Gallery; The British Museum; The Louvre; The Manchester Museum; The Metropolitan Museum of Art; The Museum of Fine Arts, Boston; The Oriental Institute of The University of Chicago.

Travel and research funding for this project was provided by a Charles Elliot Norton Fellowship and a Jens Aubrey Westengard Fund Dissertation Research Grant from Harvard University. A National Endowment for the Humanities (NEH) Digital Humanities Level II Start-up Grant facilitated the analytical and interpretative phase of this project. Gansell was supported by a Postdoctoral Fellowship at Emory University's Bill and Carol Fox Center for Humanistic Inquiry (2010-2011), and van de Meent was supported by a Rubicon Fellowship from the Netherlands Organisation for Scientific Research. Final revisions to this article were made at a writing retreat sponsored by the Faculty Writing Initiative of The Center for Teaching and Learning at St. John's University.

The authors would like to acknowledge Jasper Pope for his illustrations and Akiva Bamberger for his work on a prototype web-based visualization tool for this project. We would also like to thank Georgina Herrmann and the British Institute for the Study of Iraq (BISI) for permitting us to publish their photographs, and we would like to thank Nigel Tallis, curator of Assyrian and Babylonian artifacts at The British Museum, for his advice on image usage and patient assistance in sorting out object accession numbers. Finally, we would like to acknowledge and express our appreciation to Irene Winter and our two anonymous reviewers for their thorough and constructive feedback, from which this paper has significantly benefitted.